\documentclass[journal]{IEEEtran}
\usepackage[noadjust]{cite}
\usepackage{multirow}
\usepackage{graphicx}
\usepackage{epstopdf}
\usepackage{caption}
\usepackage{subcaption}
\usepackage{mathtools}
\usepackage{amsmath}
\usepackage[english]{babel}
\usepackage{setspace}

\begin{document}

\title{
A Low Complexity VLSI Architecture for Multi-Focus Image Fusion in DCT Domain}
\author{Ashutosh Mishra, Sudipta Mahapatra, Swapna Banerjee}

\maketitle
\begin{abstract}
Due to the confined focal length of optical sensors, focusing all objects in a scene with a single sensor is a difficult task. To handle such a situation, image fusion methods are used in multi-focus environment. Discrete Cosine Transform (DCT) is a widely used image compression transform, image fusion in DCT domain is an efficient method. This paper presents a low complexity approach for multi-focus image fusion and its VLSI implementation using DCT. The proposed method is evaluated using reference/ non-reference fusion measure criteria and the obtained results asserts it's effectiveness. The proposed method uses only $(N^2-1)$ addition to fuse the two $N\times N$ image block and consumes only 250 mW power at 200 MHz operating frequency. The maximum synthesized frequency on FPGA is found to be 221 MHz and consumes 42\% of FPGA resources. The proposed method consumes very less power and can process 4K resolution images at the rate of 60 frames per second which makes the hardware suitable for handheld portable devices such as camera module and wireless image sensors.   
\end{abstract}


\section{Introduction}
A blurred image carries less information than a sharp image. Due to the confined focal depth, it is hard to capture an image in which all objects/areas of the scene appear quite sharp. Only the objects of a scene that are near the focus plane or at the focus plane appear sharp. To handle such situations the images are acquired using a number of imaging sensors or with multi focus imaging. After taking images, high contrast regions from each of the acquired images are selected and fused together to create an image that is in focus everywhere. The fusion process reduces the uncertainty and redundancy from the source images. There are a number of benefits of using the fusion process including reduced uncertainty, wider temporal and spatial coverage and improved reliability \cite{ref1}. Image fusion is basically used due to the limited characteristic of image sensors. 

Visual sensor networks are crucial in monitoring and surveillance, tracking and object recognition. Now-a-days visual sensor networks provide help in solving various kind of problems related to research. There are multiple applications of multi-focus imaging such as in macro-photography, wireless image sensor network, focus stacking etc. In the case of Wireless Image Sensor Networks (WISN), it is hard to describe the critical situation precisely with a single image. A number of sensors are utilized to receive images of the same scene, and a centralized fusion centre cartels images from various sensors into a single image, the resultant image is amicable for human visual and machine processing \cite{ref2,ref3}. Then, the blended image is channelized to a higher node. 

The simplest image fusion approach is averaging in which two images are merged by taking average of their individual pixels. The other approaches are: choose maximum absolute or minimum absolute value of pixel intensity in two or more images. But, these approaches suffer from lower performance and bad visual quality. Further, these methods are integrated with different spatial and frequency domain processing methods to improve the output image quality  \cite{ref1,ref4,ref5,ref6}. Initial development of fusion takes place in spatial domain because of its simplicity and low computational complexity. But, use of spatial domain processes result in many undesirable effects such as blurring, low contrast and other spatial degradation. These undesirable effects imposed by spatial domain algorithms can be handled in frequency domain. So, the focus is shifting to multi-scale transforms and its various types of applications. Various fusion algorithms for multi-focus images are available using multi-scale transform  \cite{ref5,ref6,ref7,ref8,ref9,ref10,ref11,ref12}. 

The most used multi-scale transform is wavelet transform. There are a number of algorithms for image fusion that use wavelet transform \cite{ref5,ref6}. In \cite{ref7} a statistical sharpness measure is used on wavelet coefficients to perform adaptive image fusion in the wavelet domain. Instead of using DWT to decompose images into the frequency domain, Discrete Stationary Wavelet transform (DSWT) is used to overcome the lack of translation invariance of the DWT \cite{ref5,ref6}. As the wavelets do not represent long edges well in the fused images, multi-focus image fusion is performed by combining both the wavelet and curvelet transforms to improve the quality \cite{ref8}. Methods based on multi-scale transforms such as non-subsampled contourlet transform (NSCT), shift invariant discrete wavelet transform (SIDWT), and discrete wavelet transform (DWT) are popular \cite{ref5,ref6,ref9,ref10}. However, a majority of the image blending methods based on multi-scale transform are complex and time-consuming, which limit their applications for wireless visual sensor networks equipped with constrained resources.

Compliance with the image compression standards provide greater flexibility and portability to image fusion processes and algorithms. The most widely used image compression standards are JPEG and JPEG2000. In spite of having greater advantages over JPEG, JPEG2000 is not used in many in applications because of its complexity. In JPEG Discrete cosine transform (DCT) is used to compress the images. So, the DCT domain image fusion will be efficient and less time consuming. If the raw images are fused and compressed at source end (where images are generated), the compressed images take less bandwidth and less space in transmission and storage for further processing. In this paper we propose a method based on contrast measurement in the DCT domain with a reduced computational complexity. The proposed method has equal/better visual and quantitative performance as that of methods proposed in \cite{ref14}, \cite{ref3} and \cite{ref2}.

The remaining paper is arranged as follows: Section II presents the brief discussion on DCT domain fusion concept. Section III outlines the proposed approach of image fusion. Section IV explains the hardware architecture. Section V analyses the experimental results, followed by conclusions in Section VI.

\section{Prior Work and Problem Statement}
This section presents some of the most widely used concepts and algorithms based on DCT domain processing. In all of the DCT based fusion methods the maximum or average rule for fusion is applied as per the different contrast measurement criteria. Jinshan Tang \cite{ref11} proposed two methods namely; DCT+Contrast and DCT+Average. In DCT+Average, the DCT coefficient of the image is averaged to form a new fused image, but due to the averaging of the DCT coefficients the output image gets blurred. This method is not preferred due to its poor image quality. The DCT+contrast method utilizes the concept of contrast assessment presented by Tang et al. \cite{ref12}. In this method, each of the blocks is partitioned into fifteen bands of frequency  as shown in Figure 1. The sharpness of each of the coefficients in a frequency band is given as: 
\begin{equation}
{{C}_{i,j}}{\rm{  =  }}\frac{{{{\rm{d}}_{i,j}}}}{{\sum\limits_{k = 0}^{n - 1} {{E_K}} }}
\end{equation}
where ${E_K}$ is the average amplitude over the spectral band. Both the methods are criticised because of their poor performance on different evaluation criteria \cite{ref2,ref3}. Haghighat et al. \cite{ref3} proposed another method based on variance in DCT domain, namely DCT+Variance. Since variance is a better criterion than mean to estimate the contrast in the image. This method leads to a better performance than DCT+Contrast and DCT+avarage. The variance in DCT domain is given as:  
\begin{equation}
{\sigma ^2} = \sum\limits_{i = 0}^{N - 1} {\sum\limits_{j = 0}^{N - 1} {\frac{{{d^2}(i,j)}}{{{N^2}}}} }  - d(0,0)
\end{equation}
\begin{figure}
\centering
\includegraphics[scale=.5]{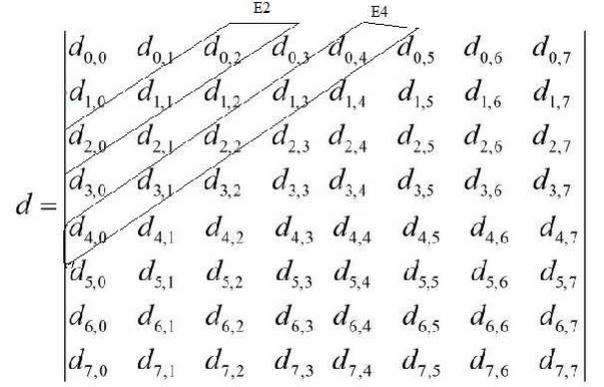}
\caption{Frequency Band in DCT+Contrast}
\end{figure}

\begin{figure*}
  \begin{subfigure}{1\textwidth}
  \centering
    \includegraphics[scale=.7]{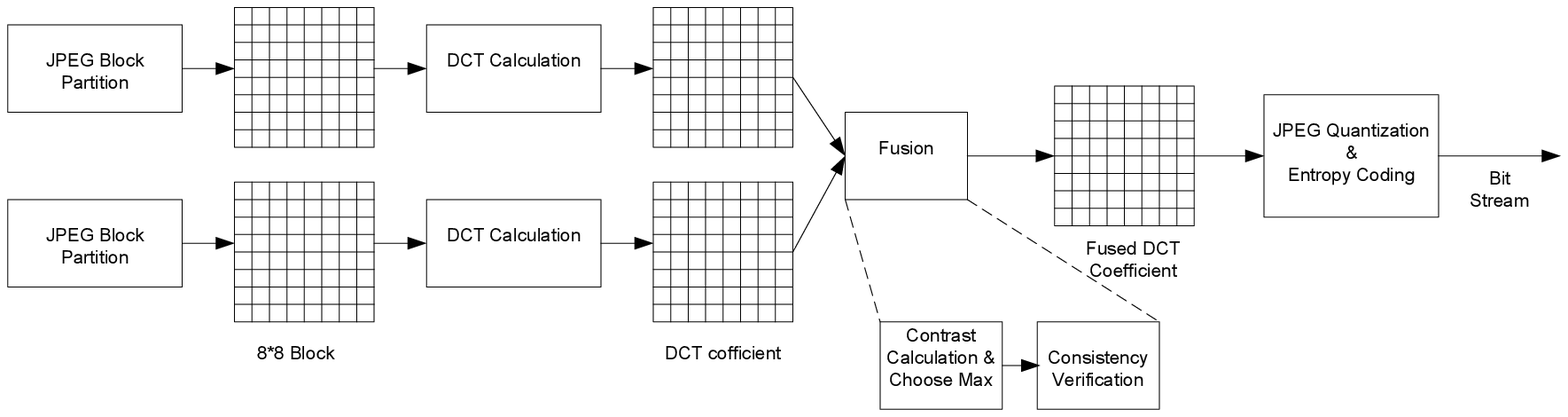}
    \caption{}
  \end{subfigure}
 \begin{subfigure}{1\textwidth}
   \centering
   \includegraphics[scale=1]{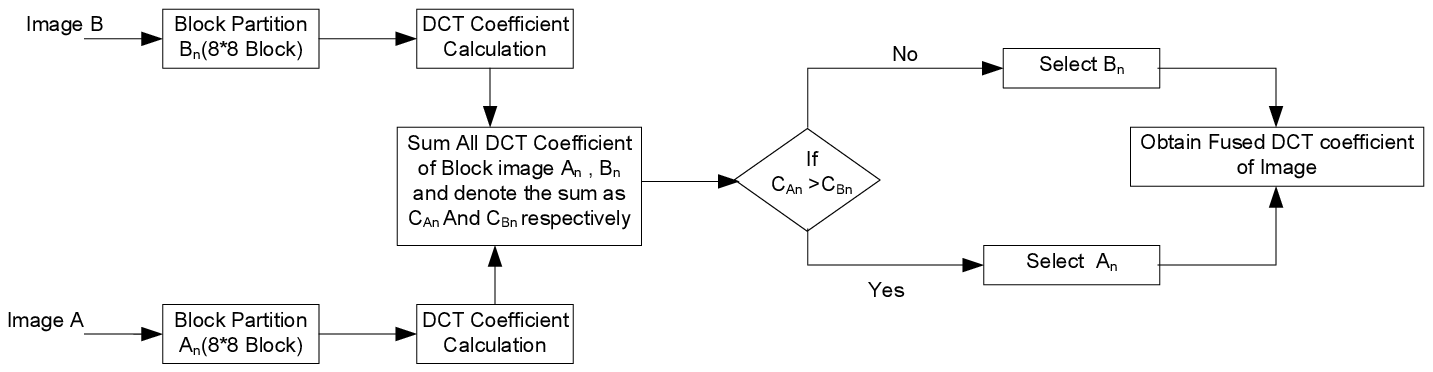}
   \caption{}
 \end{subfigure}
\caption{(a) Multi-Focus Image Fusion in JPEG Format \cite{ref2,ref3,ref11,ref14} (b) Flowchart of Multi-Focus Image Fusion}
\label{fig:architecture}
\end{figure*}

However, experimental results show that the use of variance results in a lower performance compared to that of other focus measures \cite{ref13}. The variance calculation is computationally complex because it involves floating point multiplication and division. So, it may not be suitable for resource constrained environments. The DCT+Variance method is also criticised due to its computational intensive nature. To overcome the computational intensive nature, Phamila et al. \cite{ref2} have proposed another method based on number of high valued AC coefficient in a DCT Block. Though his method reduces the overall computational complexity and energy requirement, but number of high value AC coefficient is not a good criterion to identify the maximum variance block in an image. 

Cao et al. \cite{ref14} proposed a method  named as DCT+SF. The spatial frequency calculation in DCT domain is given as : 
\begin{equation}
S{F^2} = \frac{1}{{8 \times 8}}\sum\limits_{i = 0}^{N - 1} {\sum\limits_{j = 0}^{N - 1} {E(i,j) \times d{{(i,j)}^2}} } 
\end{equation} 

Where \textit{E(i,j)} is the spatial frequency operator in the DCT domain. Calculating spatial frequency in DCT domain is computationally more complex than DCT+Variance. Due to small size of sensors, various kinds of design constraints are imposed by wireless images sensor network (WISN) on designers such as computational complexity, computational power, communication channels, image compression, and power consumption.

For resource constrained environments such as WISN, space application or any remote sensing application, the fusion method should exhibit following properties :
1) it should be computationally less complex,
2) it should consume less power,
3) Should be easier to implement on different platforms such as hardware, software or DSP platforms.
 
\section{Proposed Method}
\subsection{Contrast Measurement Criteria}
In the DCT domain coefficient \textit{d(0,0)} represents the DC value, also know as the mean value of the image block. The rest of the DCT coefficients (AC coefficients) represent the high frequency information of the image block. Determining the spatial frequency through the AC components of DCT is a computationally intensive task and also not suitable for resource constrained real-time applications such as a WISN. The coefficients of DCT itself represent spatial frequency content in given frequency band. As DCT is the part of the Discrete Fourier Transform (DFT), the definition of contrast measurement in DFT can also be utilised for DCT. The issue of contrast of complex scenes at different spatial frequencies can be addressed as follows \cite{ref15}: 

\begin{equation}
   C = \sum\limits_{k = 0}^{k = N - 1} {\sum\limits_{l = 0}^{N - 1} {} } \frac{{2F(k,l)}}{{DC}}
   \label{sch:eq2}
   \end{equation}   
where C represents the sharpness of image in the DFT domain, DC is zero frequency content and $F(k,l)$ is the DFT coefficient. Since, the DC component in the DFT domain has the maximum contribution towards the image brightness, it can be neglected in computation of contrast in DCT domain. Hence, Equation (\ref{sch:eq2}) can be modified for the DCT domain  as :
\begin{equation}
   C = \sum\limits_{k = 0}^{k = N - 1} {\sum\limits_{l = 0}^{N - 1} {} } {{d(k,l)}}-DC
   \label{sch:eq3}
   \end{equation} 
   
So, instead of computing the spatial frequency, variance or maximum number of high valued AC coefficients in a block using all the transformed AC coefficients, the proposed method adds all of the DCT coefficients to  measure the contrast of the DCT block. Thus computation is drastically reduced and the contribution of individual coefficient is also taken care of. 
   
The schematic block diagram of multi-focus image fusion in DCT domain is shown in Figure 2(a). Only the images A and B are considered for simplicity; but, the procedure can be applied to fuse more than two source images. The flow chart of the proposed algorithm is shown in Figure 2(b). 
\subsection{Algorithm}
The blending process comprises the following steps:\\

1. Take the source images A and B, divide these into $8\times8$ blocks. Denote the blocks of the two images as ${A_{n}}$ and $B_{n}$.\\

2. Calculate the DCT of the blocks and add the DCT coefficients of each of the blocks according to Equation (\ref{sch:eq3}) and denote the sums as ${C_{{A_n}}}$ and ${C_{{B_n}}}$.\\

3. Compare ${C_{{A_n}}}$ and ${C_{{B_n}}}$ and depending on which of these two is higher, select either A or B. Based on the selection prepare the decision map $W_n$ as
\begin{equation}
{W_{n}} = \left\{ \begin{array}{l}
+1{\rm{\;if\;}}{{\rm{C}}_{{A_n}}}{\rm{ > }}{{\rm{C}}_{{B_n}}}\\
{\rm{ - 1   \;if\; }}{{\rm{C}}_{{A_n}}} <= {{\rm{C}}_{{B_n}}}
\end{array} \right.
\end{equation}
4.Apply consistency verification to mend the visual quality of the output image and use a majority filter to redefine the decision map; then, construct the DCT fused image based on the new decision map as\\
\begin{equation}
F = \left\{ \begin{array}{l}
DCT\;{\rm{ of }}\;{A_n}{\rm{ if  }}\;{R_n} > 0\\
DCT\;{\rm{ of }}\;{B_n}{\rm{ if  }}\;{R_n} <  = 0
\end{array} \right.
\end{equation}
5.The standardised quantization table is used to quantise the obtained DCT coefficients in the JPEG coder and then the output bit stream is entropy coded.\\ 

The proposed method compares the absolute amplitude, so, it is named as DCT+Amp\_max in further section.  
\section{Hardware Architecture}
The block diagram of the proposed multi-focus image fusion (DCT+Amp\_max) architecture is shown in figure \ref{fig:block} (a). The proposed architecture uses an $8\times 8$ DCT. The fused DCT coefficients can be further compressed by a JPEG encoder or can be converted back to image by taking inverse DCT (IDCT) of the coefficients. IDCT Architecture is same as the DCT architecture with slight changes in coefficients.  The block wise description is as follows:
\begin{figure*}
  \begin{subfigure}{1\textwidth}
  \centering
    \includegraphics[scale=1.4]{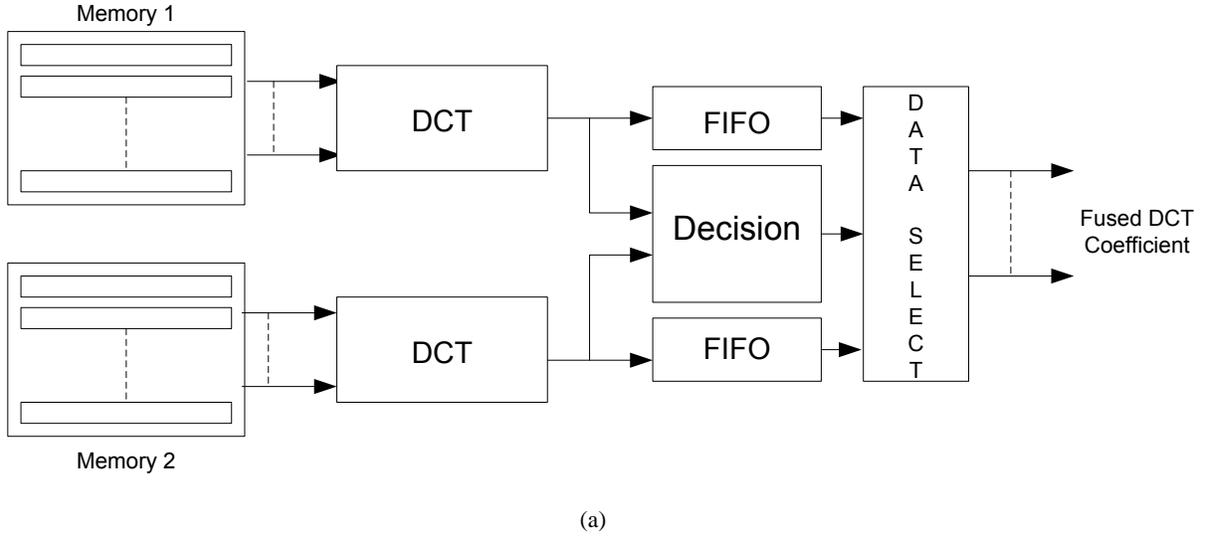}
    \caption{}
  \end{subfigure}
 \begin{subfigure}{1\textwidth}
   \centering
   \includegraphics[scale=1.4]{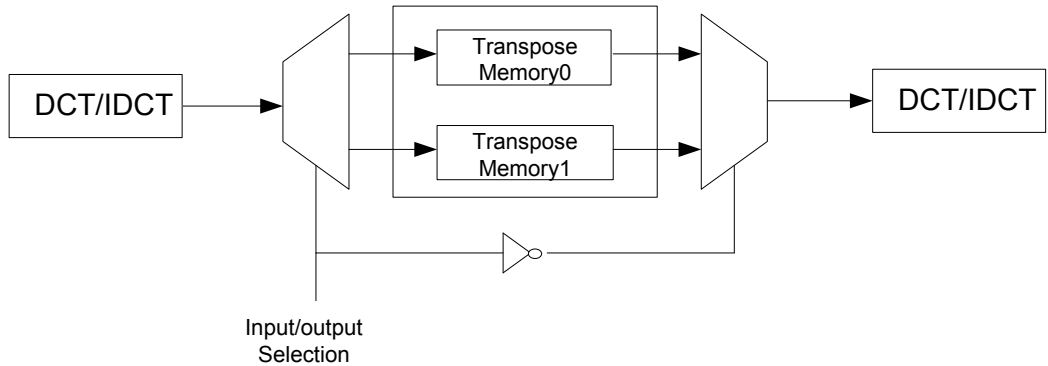}
   \caption{}
 \end{subfigure}
\caption{Hardware architecture. (a) Multi-Focus Image Fusion Block Diagram (b) DCT/IDCT Block Architecture }
\label{fig:block}
\end{figure*}

\subsection{Memory}
The presented architecture uses an $8\times 8$ DCT. It is required to configure memory in such a manner that 8 pixels per clock can be accessed, which is not possible with single memory module. Either we need to choose 4 dual port memory modules or 8 single port memory modules. In this architecture we have used 8 single port memory modules to get 8 pixels/clock cycle.
\subsection{DCT/IDCT}
Since only high frequency AC coefficients of the DCT are required to make the decision for fusion, so, an improved DCT with 35 bit (1 sign bit, 10 integer bit, 24 fractional bit) precision is synthesised. The DCT architecture is synthesised using two 1D-DCT's with an S-RAM based memory transposed architecture \cite{ref22}. The DCT architecture is shown in figure \ref{fig:block} (b). The IDCT architecture is same as the DCT architecture. Both the architecture working at same frequency. The presented architecture can be modified easily for other high speed compression applications.  
\begin{figure*}
  \begin{subfigure}{.5\textwidth}
  \centering
    \includegraphics[scale=.96]{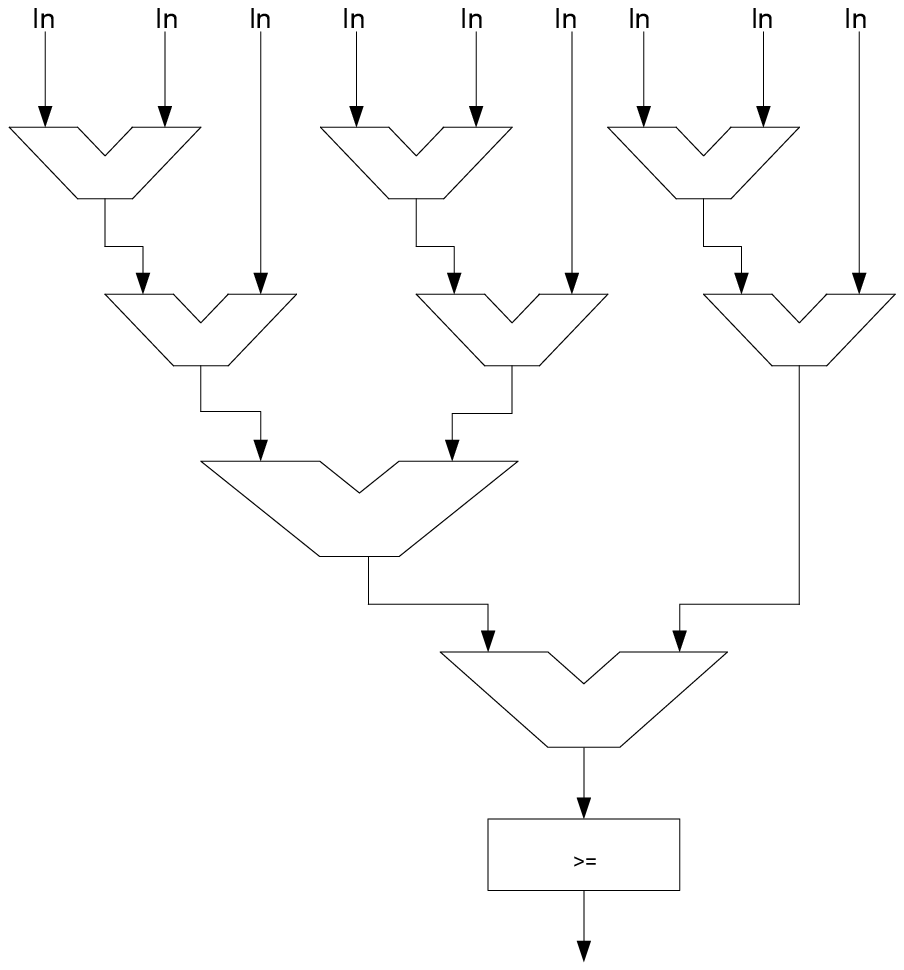}
    \caption{}
  \end{subfigure}
 \begin{subfigure}{0.5\textwidth}
   \centering
   \includegraphics[scale=1]{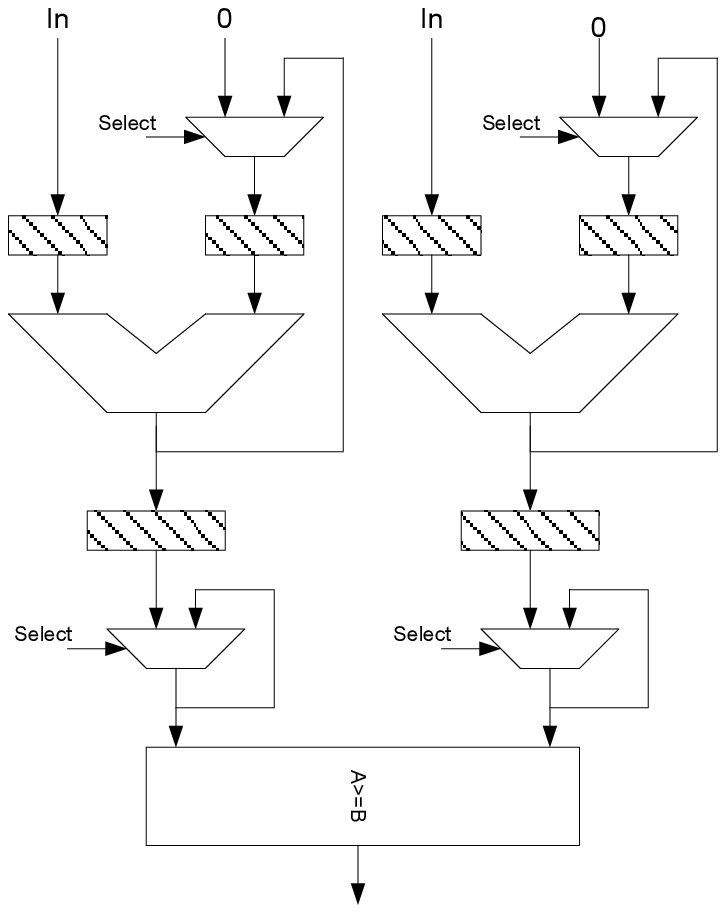}
   \caption{}
 \end{subfigure}

%
\caption{Hardware architecture for decision process. (a) Majority Filter architecture (b) Decision Block Architecture }
\label{fig:architecture1}
\end{figure*}
\subsection{Decision Block}
This block is used to select one out of two $8 \times 8$ DCT blocks. This block sums up all absolute AC coefficients of each DCT block and compares the two sums. Accordingly, it generates an output of zero or one. Simultaneously, it also prepares the decision map for consistency verification. Majority filter is applied on decision map and using newly constructed decision map, the DCT coefficient are selected according to Equation 7.  The architecture of majority filter and decision block is shown in Figure \ref{fig:architecture1}.
\subsection{First In First Out(FIFO)}
The DCT coefficient are required for fusion. The FIFO architecture helps to preserves a DCT coefficient till the final selection of the coefficients. The present FIFO is synthesised using S-Ram based memory.
\subsection{Data Select}
Based on output of Decision block the data select block selects one set of the DCT coefficients out of two sets of DCT coefficients and provides the final fused coefficient.  

\begin{figure*} 
\centering
  \includegraphics[scale=.9]{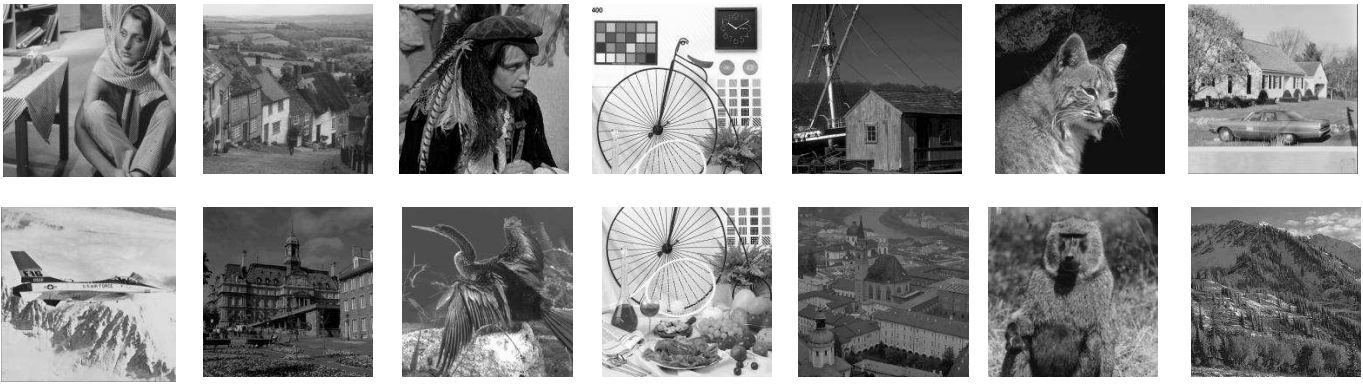}
  \caption{Test Images}\label{fig:animals}
\end{figure*} 

\begin{figure*}
        \centering
         \includegraphics[scale=.9]{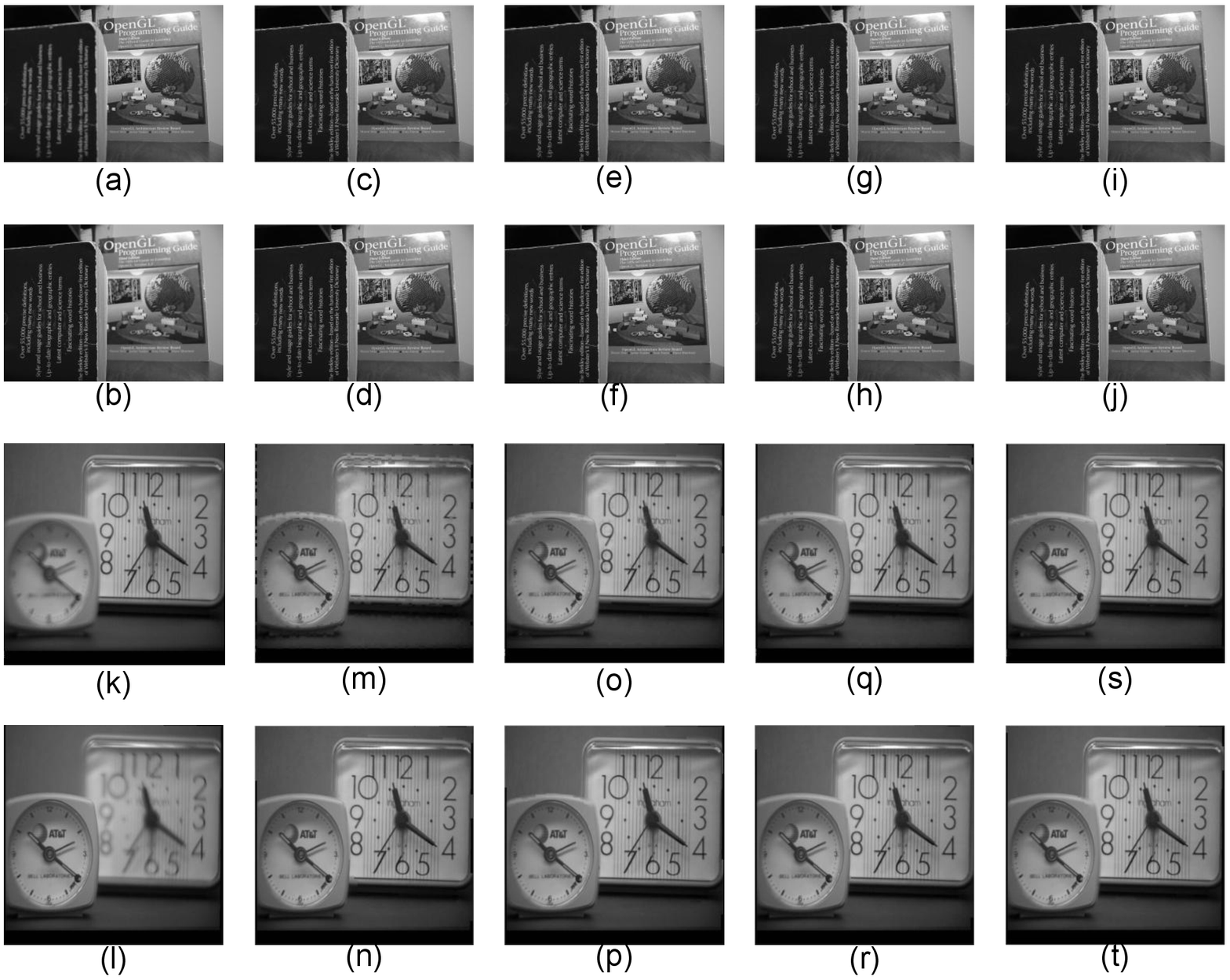}
        \caption{Image ``Clock'', ``Book'' and the fusion results. (a) \& (k) The first image with focus on the right. (b) \& (l) The second image with focus on the left. (c) \& (m) DCT$+$AC\_max. (d) \& (n) DCT$+$AC\_max$+$Cv. (e) \& (o) DCT$+$Variance. (f) \& (p) DCT$+$Variance$+$Cv. (g) \& (q) DCT$+$SF. (h) \& (r) DCT$+$SF$+$Cv. (i) \& (s) DCT+Amp\_max. (j) \& (t)DCT+Amp\_max$+$Cv.}
        \label{fig:Clock}
\end{figure*}

\begin{figure*}
        \centering
        \includegraphics[scale=.9]{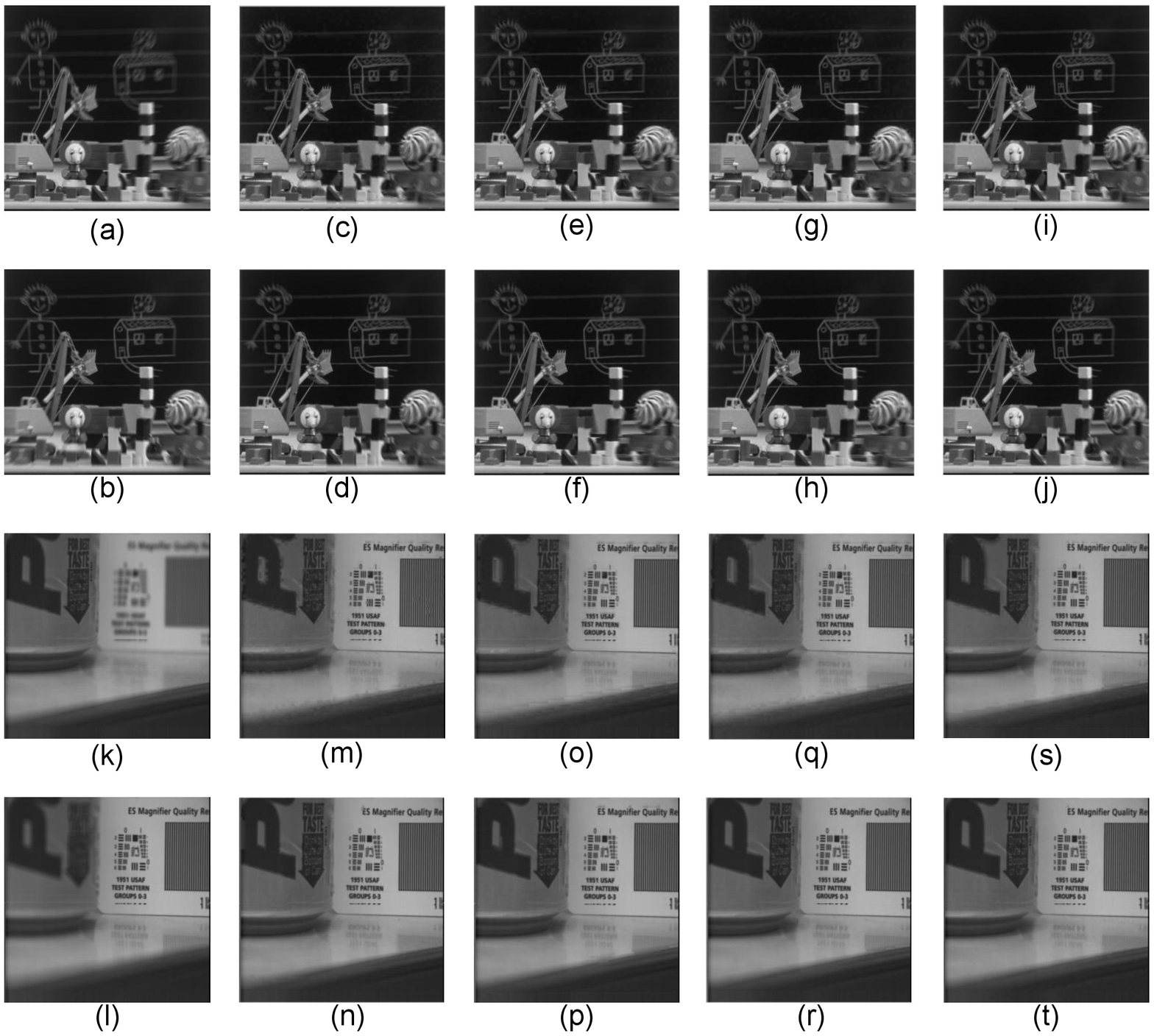}
        \caption{Image “Toy'', ``Pepsi'' and the fusion results. (a) \& (k) The first image with focus on the right. (b) \& (l) The second image with focus on the left. (c) \& (m) DCT$+$AC\_max. (d) \& (n) DCT$+$AC\_max$+$Cv. (e) \& (o) DCT$+$Variance. (f) \& (p) DCT$+$Variance$+$Cv. (g) \& (q) DCT$+$SF. (h) \& (r) DCT$+$SF$+$Cv. (i) \& (s) DCT+Amp\_max. (j) \& (t) DCT+Amp\_max$+$Cv.}
        \label{fig:pepsi2}
\end{figure*}

\section{Results and Discussion}
\subsection{Performance Analysis}
The DCT+Amp\_max method is applied to a set of reference and non-reference images and its performance is evaluated on a standard PC (Intel Core i5 Processor, 3.10 GHz, 4 GB RAM). Extensive evaluation is carried out on different images. Some of the images are shown in Figure 5 \cite{ref20,ref21}. In this section, we compare the performance of our method with that of the existing image fusion methods in the DCT domain, like DCT+Ac-max, DCT+Ac-max+Cv, DCT+Variance, DCT+Variance+Cv and DCT+SF, DCT+SF+Cv.

In the first set of experiments, the proposed method namely, DCT+Amp\_max is evaluated by fusing 50 pairs of blurred images which are generated using the standard gray scale images shown in Figure 5 with an averaging filter. Complementary regions of the images are blurred in every pair of image. The standard gray scale images are taken as ground truth images. Table \ref{sch:tb1} lists the final Universal Image Quality Index (UIQI) \cite{ref16}, Peak Signal to Noise Ratio (PSNR), Mean Square Error (MSE), Mutual Information(MI) and Structural Similarity Index Measure (SSIM) \cite{ref17} values of 50 experimental images. These are fusion evaluation criteria requiring reference images. The performance of the proposed method is approximately identical to the DCT+SF method and it has a better performance compared to that of the DCT+Variance, and DCT+Ac-Max fusion method.

\begin{table*}
\centering
\caption{Reference based Fusion Evaluation}
\begin{tabular}{c|c|c|c|c|c|}
\cline{2-6}
\multicolumn{1}{l|}{}                 & \multicolumn{5}{c|}{Objective evaluation}    \\ \hline
\multicolumn{1}{|c|}{Method}          & SSIM   & PSNR    & MSE     & MI     & UIQI   \\ \hline
\multicolumn{1}{|c|}{DCT+Ac\_max \cite{ref3}}     & 0.9763 & 34.9162 & 20.9634 & 3.7848 & 0.8303 \\ \hline
\multicolumn{1}{|c|}{DCT+Ac\_max+Cv \cite{ref3}}  & 0.9759 & 34.8722 & 21.1768 & 3.7763 & 0.8284 \\ \hline
\multicolumn{1}{|c|}{DCT+Varience \cite{ref2}}    & 0.979  & 35.1015 & 20.0879 & 3.9204 & 0.8319 \\ \hline
\multicolumn{1}{|c|}{DCT+Varience+Cv \cite{ref2}} & 0.9806 & 35.4844 & 18.3926 & 3.9512 & 0.8393 \\ \hline
\multicolumn{1}{|c|}{DCT+SF \cite{ref14}}          & 0.9844 & 36.2023 & 15.59 & 4.0281 & 0.8457 \\ \hline
\multicolumn{1}{|c|}{DCT+SF+Cv \cite{ref14}}       & 0.9844  & 36.2036 & 15.5856  & 4.0282  & 0.8457  \\ \hline
\multicolumn{1}{|c|}{DCT+Amp\_max (Proposed)}                    & 0.9836 & 36.1464 & 15.7923 & 4.005  & 0.8415 \\ \hline
\multicolumn{1}{|c|}{DCT+Amp\_max+Cv (Proposed)}                 & 0.9836 & 36.1498 & 15.78   & 4.005  & 0.8415 \\ \hline
\end{tabular}
\label{sch:tb1}
\end{table*}

The second set of experiments conducted on standard multi-focus images such as clock, pepsi, book and toy. The results of ``Book" , ``Clock", ``Toy" and ``Pepsi" are presented in Figure \ref{fig:Clock}, Figure \ref{fig:pepsi2} respectively. The performance assessment is done by the Petrovic metric $(Q^{AB/F})$ \cite{ref18}, feature mutual information (FMI) \cite{ref19} and spatial frequency (SF) measure. These metrics estimate the transfer of local structures from input images to the fused image. The higher the values of these metrics, the better is the quality of the fused image. The performance analyses of four well-known images ``Clock", ``Pepsi" , ``Book" and ``Toy" are shown in Tables \ref{sch:tb2} and \ref{sch:tb3} receptively. It is observed from Tables \ref{sch:tb2} and \ref{sch:tb3} that the performance of the proposed technique is superior or equivalent to that of the DCT+SF and DCT+Variance methods.

\begin{table*}
\centering
\caption{Non-Reference based Fusion Evaluation}
\begin{tabular}{c|c|c|c|c|c|c|}
\cline{2-7}
                                      & \multicolumn{6}{c|}{Subjective evaluation}                         \\ \cline{2-7} 
                                      & \multicolumn{3}{c|}{Book Image} & \multicolumn{3}{c|}{Clock Image} \\ \hline
\multicolumn{1}{|c|}{Method}          & $Q^{AB/F}$& SF        & FMI     & $Q^{AB/F}$ & SF       & FMI      \\ \hline
\multicolumn{1}{|c|}{DCT+Ac\_max \cite{ref3}}     & 0.8761    & 31.583    & 0.9136  & 0.8165     & 9.1052   & 0.9174   \\ \hline
\multicolumn{1}{|c|}{DCT+Ac\_max+Cv \cite{ref3}}  & 0.8825    & 30.8665   & 0.9168  & 0.8728     & 8.9245   & 0.9262   \\ \hline
\multicolumn{1}{|c|}{DCT+Varience \cite{ref2}}    & 0.8882    & 31.9622   & 0.9135  & 0.8841     & 9.3494   & 0.9234   \\ \hline
\multicolumn{1}{|c|}{DCT+Varience+Cv \cite{ref2}} & 0.8871    & 31.2181   & 0.9166  & 0.8728     & 8.931    & 0.9258   \\ \hline
\multicolumn{1}{|c|}{DCT+SF \cite{ref14}}          & 0.8917    & 32.1506   & 0.9151  & 0.8902     & 9.4037   & 0.9247   \\ \hline
\multicolumn{1}{|c|}{DCT+SF+Cv \cite{ref14}}       & 0.8858    & 31.1966   & 0.9166  & 0.8778     & 8.97     & 0.9262   \\ \hline
\multicolumn{1}{|c|}{DCT+Amp\_max (Proposed)}                    & 0.8913    & 32.101    & 0.9146  & 0.8887     & 9.4038   & 0.9242   \\ \hline
\multicolumn{1}{|c|}{DCT+Amp\_max+Cv (Proposed)}                 & 0.8834    & 31.2024   & 0.9165  & 0.878      & 8.9773   & 0.9262   \\ \hline
\end{tabular}
\label{sch:tb2}
\end{table*}

\begin{table*}
\centering
\caption{Non-Reference based Fusion Evaluation}
\begin{tabular}{c|c|c|c|c|c|c|}
\cline{2-7}
                                      & \multicolumn{6}{c|}{Subjective Evaluation}                            \\ \cline{2-7} 
                                      & \multicolumn{3}{c|}{Pepsi Image} & \multicolumn{3}{c|}{Toy Image}     \\ \hline
\multicolumn{1}{|c|}{Method}          & $Q^{AB/F}$& SF        & FMI      & $Q^{AB/F}$ & SF          & FMI    \\ \hline
\multicolumn{1}{|c|}{DCT+Ac\_max \cite{ref3}}     & 0.9159    & 13.1706   & 0.9231   & 0.8489    & 10.7993     & 0.9093 \\ \hline
\multicolumn{1}{|c|}{DCT+Ac\_max+CV \cite{ref3}}  & 0.9372    & 13.0021   & 0.9246   & 0.8468    & 10.2325     & 0.9153 \\ \hline
\multicolumn{1}{|c|}{DCT+Varience \cite{ref2}}    & 0.9321    & 13.2962   & 0.9238   & 0.8757    & 11.0290     & 0.9112 \\ \hline
\multicolumn{1}{|c|}{DCT+Varience+Cv \cite{ref2}} & 0.9385    & 13.0457   & 0.9246   & 0.8637    & 10.4667     & 0.9165 \\ \hline
\multicolumn{1}{|c|}{DCT+SF \cite{ref14}}          & 0.9351    & 13.3753   & 0.9243   & 0.874     & 11.0961     & 0.9106 \\ \hline
\multicolumn{1}{|c|}{DCT+SF+Cv \cite{ref14}}       & 0.9375    & 13.0162   & 0.9246   & 0.8564    & 10.3320     & 0.9156 \\ \hline
\multicolumn{1}{|c|}{DCT+Amp\_max (proposed)}                    & 0.935     & 13.2863   & 0.9242   & 0.8759    & 11.1030     & 0.9114 \\ \hline
\multicolumn{1}{|c|}{DCT+Amp\_max+Cv (Proposed)}                 & 0.9363    & 13.0032   & 0.9246   & 0.863     & 10.3620     & 0.9162 \\ \hline
\end{tabular}
\label{sch:tb3}
\end{table*}

\subsection{Hardware Implementation \& Computational Complexity}
The hardware architecture of the proposed method is implemented on Virtex4-xc4vlx200-11f1513 using the Xilinx tool. The hardware utilization and maximum synthesised frequency along with the power and area constraint on 90nm technology is reported in Table \ref{sch:tb4}. The fusion result from hardware implementation is also compared with MATLAB results with 24 bit (fractional part) precision. Both the results match each other with good accuracy. The latency in hardware is 70 clock cycles. 
\begin{table*}
\caption{ Hardware utilization, Power and Area in proposed algorithm }
\begin{center}
\begin{tabular}{|c|c|c|c|c|c|c|} \hline
Logic        & Used  & Available   & \%         & Maximum & Power   & Area   \\
Utilization& &       & Utilization & Frequency  & @200MHz  & (Gate count)  \\ \hline
Slices       & 38094 & 89088       & 42         &          &        &\\ \cline{1-4}
Slice FF pair& 66878 & 178176      & 37         & 221 MHz  & 250 mW & 846K \\ \cline{1-4}
4 input LUTs &41493  & 178176      & 23         &          &        & \\  \hline
\end{tabular}
\end{center}
\label{sch:tb4}
\end{table*}

The different algorithms of the DCT domain fusion have different computational complexities. Apart from the DCT calculation, several other multiplications and additions are required for DCT domain image fusion. Table \ref{sch:tb5} summarises the number of multiplications and additions required for $N\times N$ image block. The method DCT+Ac-max uses conditional increment, which leads to increase in hardware complexity than the proposed DCT+Amp\_max method. The proposed method is computationally less complex than the other methods without sacrificing the performance.   
\begin{table}
\caption{ Number Multiplication and addition required for $N\times N$ block}
\begin{center}
\begin{tabular}{|c|c|c|c|}
\hline
Method & Multiplication & Addition & Comparison \\ \hline
DCT+Ac\_max&\- & conditional &1 \\ 
  & &increment& \\ \hline
DCT+SF& $2(n^{2})$ & $(n^{2}-1)$ & 1\\ \hline
DCT+Varience & $(n^{2})$ &$(n^{2}-1)$ &1 \\ \hline
Proposed &0	&$(n^{2}-1)$&1	\\ \hline
\end{tabular}
\end{center}
\label{sch:tb5}
\end{table}
\section{Conclusions}
The different design constraints such as power consumption, computational power, image compression, and communication channels for WISN/camera system can be handled by low complexity and high performance algorithms. In terms of quantitative evaluation, the performance of the proposed method is approximately identical to most recent DCT+SF method at reduced computational complexity. The method requires only $(N^2-1)$ addition for the fusion of $N\times N$ size of image. 

The hardware implementation of the proposed method consumes less power (250 mW at 200 MHz).  The fusion approach adopted in the proposed method does not require any floating point arithmetic operations like variance, mean and spatial frequency calculation, which makes it suitable for resource constrained battery powered sensors for energy efficient fusion and subsequent compression. The proposed architecture is also suitable for high speed applications, portable and hand-held devices because it can process more than 60 frames per second with the resolution of 4K and acquired very less area.

\end{document}